%% file: main.tex
\documentclass{ecai}
\usepackage{times}
\usepackage{graphicx}
\usepackage{latexsym}
\ecaisubmission
\usepackage{amsmath}
\usepackage{upgreek}
\usepackage{amssymb}
\usepackage{stmaryrd}
\usepackage{units}
\usepackage{bm}
\usepackage{textcomp}
\usepackage{xcolor}
\usepackage{balance}
\newcommand{\ch}{\textcolor{black}}
\newcommand{\cam}{\textcolor{black}}
\begin{document}

\title{Forecaster: A Graph Transformer for Forecasting \\Spatial and Time-Dependent Data}

\author{Yang Li \and Jos\'e M. F. Moura\institute{Carnegie Mellon University,
USA\hfill\break 
Email: \{yangli1, moura\}@andrew.cmu.edu}}

\maketitle
\bibliographystyle{ecai}
\input{Abstract}
\input{Introduction}

\input{Methodology}
\input{Evaluation}
\input{RelatedWork}
\input{Conclusion}
\bibliography{reference}
\input{Appendix}
\end{document}

%% file: Abstract.tex
\begin{abstract}
Spatial and time-dependent data is of interest in many applications.
This task is difficult due to its complex spatial dependency, long-range
temporal dependency, data non-stationarity, and data heterogeneity.
To address these challenges, we propose Forecaster, a graph Transformer
architecture. Specifically, we start by learning the structure of
the graph that parsimoniously represents the spatial dependency between the data at different locations. Based on the topology of the graph, we sparsify the
Transformer to account for the strength of spatial dependency, long-range
temporal dependency, data non-stationarity, and data heterogeneity.
We evaluate Forecaster in the problem of forecasting
taxi ride-hailing demand and show that our proposed architecture significantly
outperforms the state-of-the-art baselines.
\end{abstract}

%% file: Introduction.tex
\section{Introduction}

Spatial and time-dependent data describe the evolution of signals
(i.e., the values of attributes) at multiple spatial locations across
time \cite{Spatial_Time_Series_Def_2003,Spatial_Time_Series_Def_2019}.
It occurs in many domains, including economics \cite{STS_Example_Economy},
global trade \cite{STS_Example_Global_Trade}, environment studies
\cite{STS_Example_Environment_Studies}, public health \cite{STS_Example_Public_Health},
or traffic networks \cite{DCRNN_ICLR2018} to name a few. For example, the gross
domestic product (GDP) of different countries in the past century,
the daily temperature measurements of different cities for the last
decade, and the hourly taxi ride-hailing demand at various urban locations
in the recent year are all spatial and time-dependent data. Forecasting such data allows
to proactively allocate resources and take actions to improve the
efficiency of society and the quality of life.

However, forecasting spatial and time-dependent data is challenging
\textemdash{} they exhibit complex spatial dependency, long-range
temporal dependency, heterogeneity, and non-stationarity. Take the
spatial and time-dependent data in a traffic network as an example.
The data at a location (e.g., taxi ride-hailing
demand) may correlate more with the data at a geometrically remote
location than a nearby location \cite{DCRNN_ICLR2018}, \cam{exhibiting complex spatial dependency}. Also, the data at a time instant may \cam{be similar to} the data at a recent time instant, say an hour ago, but may \cam{also} highly correlate with the data a day ago or even a week ago, showing strong
long-range temporal dependency. Additionally, the spatial and time-dependent
data may be influenced by many other relevant factors (e.g., weather
influences taxi demand). These factors
are relevant information, shall be taken into account. In other words, in this paper, we propose to perform forecasting
with heterogeneous sources of data at different spatial and time scales and including auxiliary information of a different nature or modality. Further, the data may be non-stationary due to unexpected incidents
or traffic accidents \cite{DCRNN_ICLR2018}. This non-stationarity
makes the conventional time series forecasting methods such as \ch{auto-regressive
integrated moving average (ARIMA) and vector autoregression (VAR)}, which usually
rely on stationarity, inappropriate for
accurate forecasting with spatial and time-dependent data \cite{DCRNN_ICLR2018,Spatio_Temporal_ICDM_2017}.

Recently, deep learning models have been proposed for forecasting for spatial
and time-dependent data \cite{DCRNN_ICLR2018,DMVSTNET_AAAI2018,jain2016structural,STGCN_IJCAI2018,STMGCN_AAAI2019,STResNet_AAAI17,Yao_AAAI2019,Spatio_Temporal_ICDM_2017}.
To deal with spatial dependency, most of these models either use pre-defined
distance/similarity metrics or other prior knowledge like adjacency
matrices of traffic networks to determine dependency among locations. Then,
they often use a (standard or graph) convolutional neural network
(CNN) to better characterize the spatial dependency between these locations.
These ad-hoc methods may lead to errors in \cam{some}  cases. 
For example, the locations
that are considered as \cam{being} dependent \cam{(independent)} may actually be independent \cam{(dependent)} in practice.
\cam{As a result, these models may encode the data at a location by considering the data at independent locations and neglecting the data at dependent locations,
leading to inaccurate encoding.}
Regarding temporal dependency, most of these models use recurrent neural
networks (RNN), CNN, or their variants to capture the data long-range temporal
dependency and non-stationarity. But it is well documented that
these networks may fail to capture temporal dependency between
distant time epochs \cite{long_range_dependency_rnn,Transformer}.

To tackle these challenges, we propose \textit{Forecaster}, a new deep learning
architecture for forecasting spatial and time-dependent data.
Our architecture consists of two parts. First, we use the theory of
Gaussian Markov random fields \cite{Gaussian_Markov_Book} to learn the structure of the graph that parsimoniously represents the spatial dependency between the locations (we call such graph a \textit{dependency graph}). Gaussian Markov random fields
model spatial and time-dependent data as a multivariant Gaussian distribution
over the spatial locations. We then estimate the precision matrix
of the distribution \cite{Graphical_Lasso}.\footnote{The approach to estimate the precision matrix of a Gaussian Markov random field (i.e., graphical lasso) can also be used with non-Gaussian distributions \cite{LogDet}.} The precision matrix provides the graph structure with each node representing a location and each edge representing the dependency between two locations. This contrasts prior work on forecasting \textemdash{} \textit{we learn from the data its spatial dependency}. Second, we integrate the dependency graph in the architecture of the Transformer \cite{Transformer} for forecasting spatial and time-dependent data. The
Transformer and its extensions \cite{Transformer,Transformer_XL,XLNet,Bert,OpenAI_Transformer}
have been shown to significantly outperform RNN and CNN in
NLP tasks, as they capture relations among data at distant positions, significantly improving the learning
of long-range temporal dependency \cite{Transformer}. In our Forecaster, in order to better capture the spatial dependency, we associate each neuron
in different layers with a spatial location. Then,
we sparsify the Transformer based on the dependency graph:
if two locations are not connected \ch{in the graph}, we prune the connection between their
associated neurons. In this way, the state encoding for each location
is only impacted by its own state encoding and encodings
for other dependent locations. Moreover, pruning the unnecessary connections in the Transformer avoids overfitting.

To evaluate the effectiveness of our proposed architecture, we apply
it to the task of forecasting taxi ride-hailing demand in New York
City \cite{NYC_Taxi}. We pick 996 hot locations in New York City and
forecast the hourly taxi ride-hailing demand around each location
from January 1st, 2009 to June 30th, 2016. Our architecture accounts for crucial auxiliary
information such as weather, day of the week, hour of the day, and holidays. This improves significantly the forecasting
task. Evaluation results show that our architecture reduces the root
mean square error (RMSE) and mean absolute percentage error (MAPE)
of the Transformer by 8.8210\% and 9.6192\%,
respectively, and also show that our architecture significantly outperforms
other state-of-the-art baselines.

In this paper, we present critical innovation:\begin{itemize}\vspace{-4pt}
\item Forecaster combines the theory of Gaussian Markov random fields with
deep learning. It uses the former to find the dependency graph among locations, and this graph becomes the
basis for the deep learner forecast spatial and time-dependent data.
\item Forecaster sparsifies the architecture of the Transformer based on the dependency graph, allowing the Transformer to capture better
the spatiotemporal dependency within the data.
\item We apply Forecaster to forecasting taxi ride-hailing
demand and demonstrate the advantage of its proposed architecture
over state-of-the-art baselines.
\end{itemize}


%% file: Methodology.tex
\section{Methodology}

In this section, we introduce the proposed architecture of Forecaster.
We start by formalizing the problem of forecasting spatial and time-dependent data 
(Section \ref{subsec:Problem-Statement}). Then, we use Gaussian Markov random fields to determine the
dependency graph among data at different locations (Section \ref{subsec:GMRF}).
Based on this dependency graph, we design a sparse linear layer, which is
a fundamental building block of Forecaster (Section \ref{subsec:Building-Block}).
Finally, we present the entire architecture of Forecaster (Section \ref{subsec:Entire-Architecture}).

\subsection{\label{subsec:Problem-Statement}Problem Statement}

We define spatial and time-dependent data as a series of \textit{spatial signals}, each collecting the data at 
all spatial locations at a certain time.
For example, hourly taxi demand at a thousand locations in 2019 is a spatial and time-dependent data, while the hourly taxi demand at these locations between 8 a.m. and 9 a.m. of January 1st, 2019 is a spatial signal. The goal of our forecasting task is to predict the future spatial signals
given the historical spatial signals and historical/future
auxiliary information (e.g., weather history and forecast). We formalize forecasting
as learning a function $h\left(\cdot\right)$ that maps $T$ historical
spatial signals and $T+T'$ historical/future auxiliary information to $T'$ future spatial signals, as Equation
\eqref{eq:1}:\begin{small}
\begin{equation}
\left[\begin{array}{ccc}
\mathbf{{x}}_{t-T+1}, & \cdots, & \mathbf{{x}}_{t};\\
\mathbf{{a}}_{t-T+1}, & \cdots, & \mathbf{{a}}_{t+T'}
\end{array}\right]\overset{h\left(\cdot\right)}{\longrightarrow}\left[\begin{array}{ccc}
\mathbf{{x}}_{t+1}, & \cdots, & \mathbf{{x}}_{t+T'}\end{array}\right]\label{eq:1}
\end{equation}
\end{small}where $\mathbf{{x}}_{t}$ is the spatial signal at time $t$, $\mathbf{{x}}_{t}=\left[\begin{array}{ccc}
x_{t}^{1}, & \cdots, & x_{t}^{N}\end{array}\right]^{T}\in\mathbb{{R}}^{N}$, with $x_{t}^{i}$ the data at location $i$ at time $t$; $N$
the number of locations;  $\mathbf{{a}}_{t}$ the auxiliary information at
time $t$, $\mathbf{{a}}_{t}\in\mathbb{{R}}^{P}$, $P$ the dimension of the auxiliary
information;\footnote{For simplicity, we assume in this work that different locations share the same auxiliary information, i.e., $\mathbf{{a}}_{t}$ can impact $x_{t}^{i}$, for any $i$. However, it is easy to generalize our approach to the case where locations do not share the same auxiliary information.} and $\mathbb{R}$ is the set of the reals.

\subsection{\label{subsec:GMRF}Gaussian Markov Random Field}

We use Gaussian Markov random fields to find the dependency
graph of the data over the different spatial locations. Gaussian Markov random fields model
the spatial and time-dependent data $\mathbf{\left\{ \mathbf{{x}_{t}}\right\} }$
as a multivariant Gaussian distribution over $N$ locations,
i.e., the probability density function of the vector given by $\mathbf{{x}_{t}}$ is
\begin{equation}
f\left(\mathbf{{x}_{t}}\right)=\frac{\left|Q\right|}{\left(2\pi\right)^{\unitfrac{N}{2}}}\exp\left(-\frac{1}{2}\left(\mathbf{\mathbf{{x}}}_{t}-\mathbf{\bm{{\mu}}}\right)^{T}Q\left(\mathbf{\mathbf{{x}}}_{t}-\bm{{\mu}}\right)\right)
\end{equation}
where $\bm{{\mu}}$ and $Q$ are the expected value (mean) and precision matrix (inverse of the covariance matrix)
of the distribution.

The precision matrix characterizes the conditional dependency
between different locations \textemdash{} whether the data $x_{t}^{i}$ and $x_{t}^{j}$ at the $i^{\mathrm{{th}}}$
and $j^{\mathrm{{th}}}$ locations depend
on each other or not given the data at all the other locations
$x_{t}^{-ij}$ ($x_{t}^{-ij}=\left\{ x_{t}^{k}\:|\:k\neq i,\,j\right\} $).
We can measure the conditional dependency between locations $i$ and
$j$ through their conditional correlation coefficient $\mathrm{{Corr}}\left(x_{t}^{i},x_{t}^{j}\:|\:x_{t}^{-ij}\right)$:
\begin{equation}
\mathrm{{Corr}}\left(x_{t}^{i},x_{t}^{j}\:|\:x_{t}^{-ij}\right)=-\frac{Q_{ij}}{\sqrt{Q_{ii}Q_{jj}}}
\end{equation}
where $Q_{ij}$ is the $i^{\mathrm{{th}}}$, $j^{\mathrm{{th}}}$ entry of $Q$. In practice,
we set a threshold on $\mathrm{{Corr}}\left(x_{t}^{i},x_{t}^{j}\:|\:x_{t}^{-ij}\right)$,
and treat locations $i$ and $j$ as conditionally dependent if the
absolute value of $\mathrm{{Corr}}\left(x_{t}^{i},x_{t}^{j}\:|\:x_{t}^{-ij}\right)$
is above the threshold.

\begin{figure}[tbh]
\centering
\includegraphics[width=0.35\columnwidth]{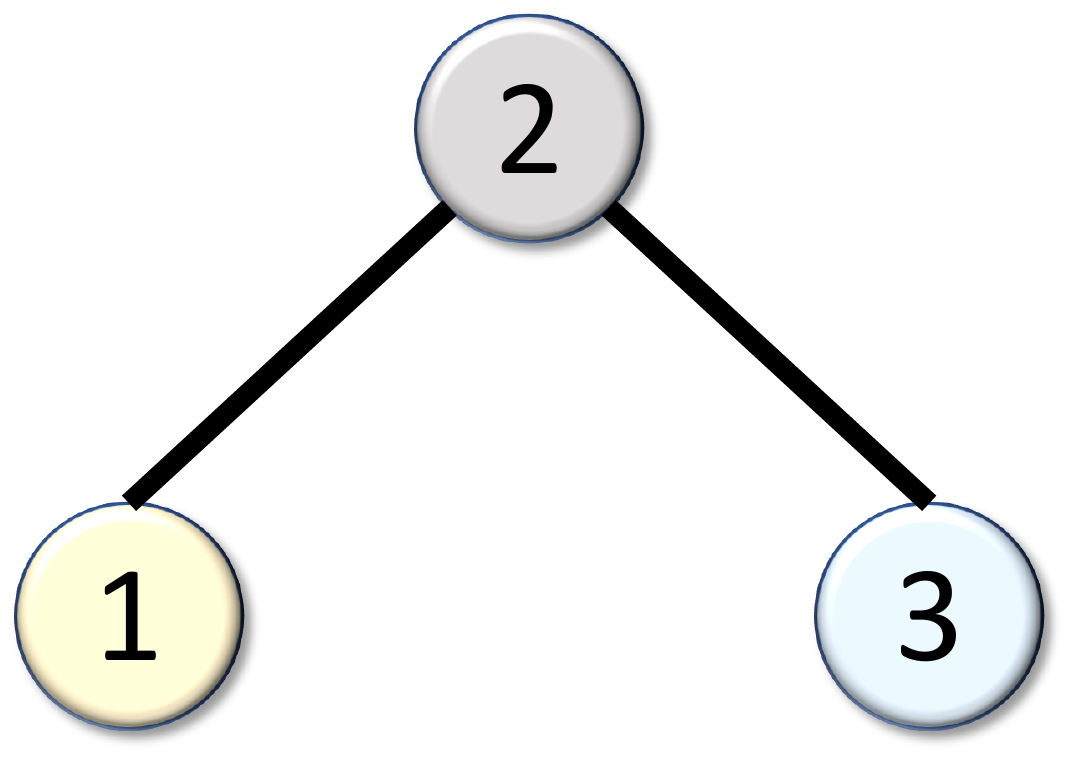}
\caption{\label{fig:Example-for-dependency}\ch{An example of a simple dependency graph.}}
\end{figure}

The non-zero entries define the structure of the \textit{dependency graph} between locations.
Figure \ref{fig:Example-for-dependency} shows an example of a dependency
graph. Locations 1 and 2 and locations 2 and 3 are conditionally dependent, while locations 1 and 3 are conditionally independent. This principle example illustrates the advantage of Gaussian Markov random field over ad-hoc
pairwise similarity metrics \textemdash{} the former leads to parsimonious (sparse) graph representations.

We estimate the precision matrix by graphical lasso \cite{Graphical_Lasso},
an L1-penalized maximum likelihood estimator: 
\begin{equation}
\begin{array}{c}
\underset{Q}{\textrm{{min}}} \enskip\mathrm{{tr}}\left(SQ\right)-\mathrm{\log}\det\left(Q\right)+\lambda\left\Vert Q\right\Vert _{1}\\
s.t.,\enskip Q\in\left\{ Q=Q^{T},\;Q\succ0\right\} 
\end{array}
\end{equation}
where $S$ is the empirical covariance matrix computed from the data:
\begin{equation}
\begin{array}{r@{}l}
S&{}=\frac{1}{M-1}\sum_{t=1}^{M}\left(\mathbf{\mathbf{{x}}}_{t}-\mathbf{\bm{{\mu}}}\right)^{T}\left(\mathbf{\mathbf{{x}}}_{t}-\bm{{\mu}}\right)\\ \\
\textrm{\ensuremath{\bm{{\mu}}}}&{}=\frac{1}{M}\sum_{t=1}^{M}\mathbf{\mathbf{{x}}}_{t}
\end{array}
\end{equation}
where $M$ is the number of time samples used to compute $S$.

\subsection{\label{subsec:Building-Block}Building Block: Sparse Linear Layer}

We use the dependency graph to sparsify
the architecture of the Transformer. This leads to the Transformer better
capturing the spatial dependency within the data. There are multiple linear
layers in the Transformer. Our sparsification on the Transformer
replaces all these linear layers by the sparse
linear layers described in this section. 

\begin{figure}[tbh]
\centering
\includegraphics[width=0.9\columnwidth]{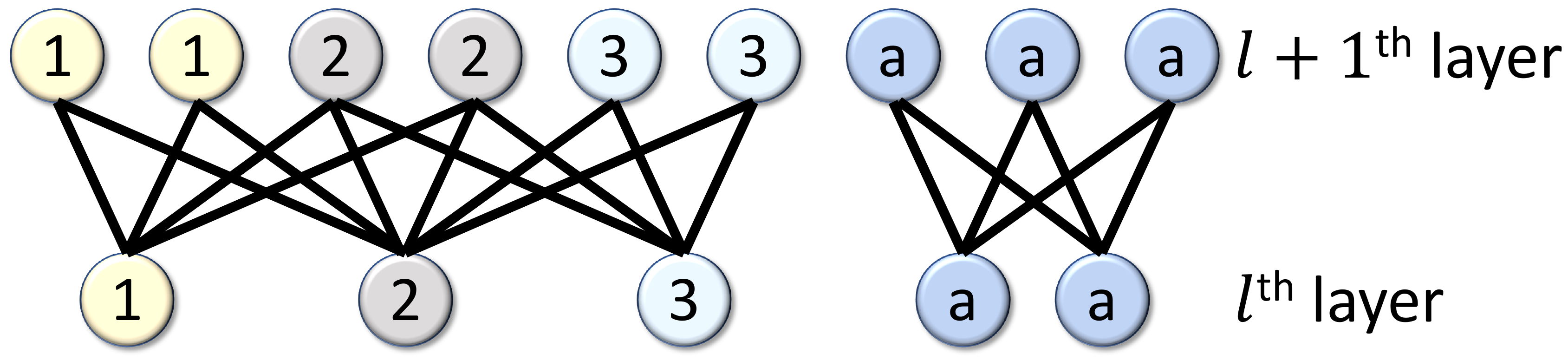}
\caption{\label{fig:sparse-linear}\ch{An example of a sparse linear layer based on
the dependency graph in Figure \ref{fig:Example-for-dependency} (neurons marked as "1", "2", "3", and "a" are for locations 1, 2, and 3, and auxiliary information, respectively).}}
\end{figure}

We use the dependency graph to build a sparse linear layer. Figure \ref{fig:sparse-linear} shows an example
(based on the dependency graph in Figure \ref{fig:Example-for-dependency}).
Suppose that initially \ch{the $l^{\mathrm{{th}}}$ layer (of five neurons) is fully connected to the $l+1^{\mathrm{{th}}}$ layer (of nine neurons).}
We assign neurons to the data at different locations \ch{(marked as "1", "2", and "3" for locations 1, 2, and 3, respectively)} and to the auxiliary information \ch{(marked as "a")}
as illustrated next. \ch{How to assign neurons is a design choice for users.}  In this example, assign one neuron to each
location and two neurons to the auxiliary information at the $l^{\mathrm{{th}}}$
layer and assign two neurons to each location and three neurons to
the auxiliary information at the $l+1^{\mathrm{{th}}}$ layer.
After assigning neurons, we prune connections based on the structure
of the dependency graph. As location\ch{s} 1 and 3 are conditionally independent, we prune
the connections between them. We also prune the connections between
the neurons associated with locations and the auxiliary information \ch{to further simplify the architecture.}\footnote{\ch{However, our architecture still allows the encodings for the data at different locations (i.e., the encoding for the spatial signal) to consider the auxiliary information through the sparse multi-head attention layers in our architecture, which we will illustrate in the Section \ref{subsec:Entire-Architecture}.}}
This way, the encoding for the data at a location is only impacted by the encodings
of itself and of its dependent locations, better capturing
the spatial dependency between locations. Moreover, pruning the unnecessary
connections between conditionally independent locations helps avoiding overfitting.

\cam{Our sparse linear layer is similar to the state-of-the-art graph convolution approaches such as GCN \cite{gcn} and TAGCN \cite{tagcn, shi2018}  \textemdash{} all of them transform the data based on the adjacency matrix of the graph.} The major difference is our sparse linear layer learns the weights for non-zero entries of the adjacency matrix (equivalent to the weights of the sparse linear layer), considering that different locations may have different strengths of dependency between each other.  

\subsection{\label{subsec:Entire-Architecture}Entire Architecture: Graph Transformer}

\begin{figure}[b]
\centering
\includegraphics[width=0.88\columnwidth]{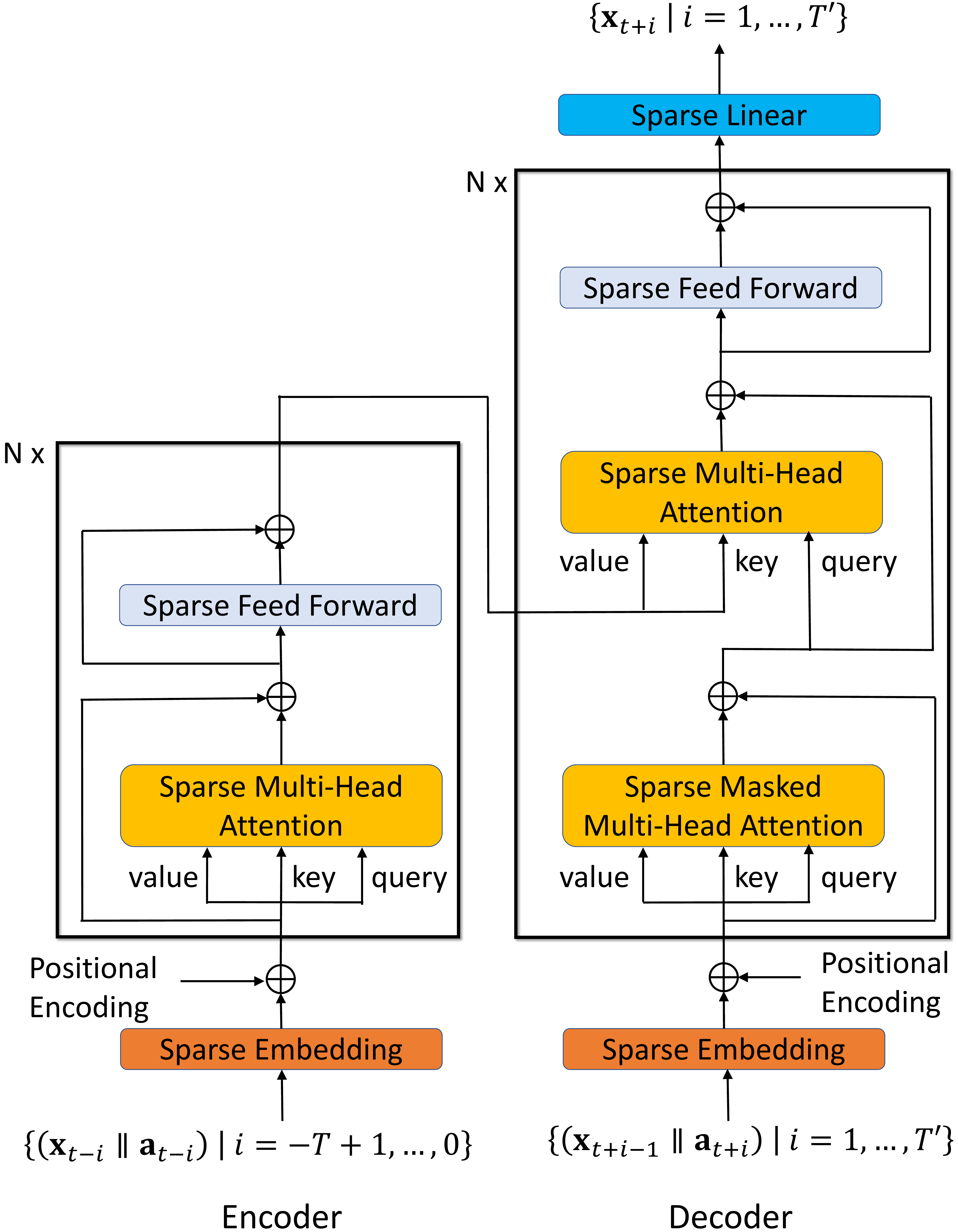}
\caption{\label{fig:Architecture-of-Forecaster}\ch{Architecture of Forecaster ($a\parallel b$ represents concatenating vector $a$ with vector
$b$).}}
\end{figure}

Forecaster adopts an architecture similar to that of the Transformer except for substituting all
the linear layers in the Transformer with our sparse linear layer designed based on the 
dependency graph.
Figure \ref{fig:Architecture-of-Forecaster} shows its architecture. Forecaster employs an encoder-decoder architecture
\cite{Seq2Seq}, \ch{which has been widely adopted in sequence generation tasks such as taxi demand forecasting \cite{DCRNN_ICLR2018} and pose prediction \cite{walker2017pose}.} 
The encoder is used to encode the historical spatial signals and historical auxiliary information; the
decoder is used to predict the future spatial signals
based on the output of the encoder and the future auxiliary information.
\ch{We omit what Forecaster shares with the Transformer (e.g., positional encoding, multi-head attention) 
and emphasize only on their differences in this section. Instead, we provide a brief introduction to multi-head attention in the appendix.} 

\subsubsection{Encoder}
At each time step in the history, we concatenate the spatial signal with its auxiliary information. This way, we
obtain a sequence where each element is a vector consisting of the spatial signal and the auxiliary information at a specific time
step. The encoder takes this sequence as input. Then, a \textit{sparse
embedding} layer (consisting of a sparse linear layer with ReLU activation)
maps each element of this sequence to the \textit{state space of the model} and outputs a new sequence. In Forecaster, except for the sparse linear layer at the end of the decoder, all the layers have the same output dimension. We term this dimension $d_{model}$ and the space with this dimension as the \textit{state space of the model}.  
After that, we add \textit{positional encoding} to the new sequence, \ch{giving temporal order information to each element of the sequence.
Next,} we let the \ch{obtained} sequence pass through $N$ stacked \textit{encoder layers} to generate the encoding of the input sequence. 
Each encoder layer consists of a \textit{sparse multi-head attention} layer and a \textit{sparse feedforward} layer.
\ch{These layers are the same multi-head attention layer and feedforward layer as in the Transformer, except that sparse linear layers, which reflect the spatial dependency between locations, to replace linear layers within them. The sparse multi-head attention layer enriches the encoding of each element with the information of other elements in the sequence, capturing the long-range temporal dependency between elements. It takes each element as a \textit{query}, as a \textit{key}, and also as a \textit{value}. 
A query is compared with other keys to obtain the similarities between an element and other elements, and then these similarities are used to weight the values
to obtain the new encoding of the element. Note each query, key, and value consists of two parts: the part for encoding the spatial signal and the part for encoding the auxiliary information \textemdash{} both impact the similarity between a query and a key. As a result, in the new encoding of each element, the part for encoding the spatial signal takes into account the auxiliary information. The sparse feedforward layer further refines the encoding of each element.}

\subsubsection{Decoder}
For each time step in the future, we concatenate its auxiliary information
with the (predicted) spatial signal one step before.
Then, we input this sequence to the decoder. The decoder first uses
a \textit{sparse embedding} layer to map each element of the sequence to the state space of the
model, \ch{adds the \textit{positional encoding},} and then passes it through $N$ stacked \textit{decoder layers} to obtain
the new encoding of each element. Finally, the decoder uses a \textit{sparse linear
layer} to project this encoding back and predict
the next spatial signal. Similar to the Transformer, the
decoder layer contains two \textit{sparse multi-head attention} layers \ch{and a \textit{sparse feedforward layer}}. The first
\ch{(masked) sparse multi-head attention layer} compares the elements in the sequence, obtaining
a new encoding for each element. 
\ch{Like the Transformer, we put a mask here such that an element is compared with only earlier elements in the sequence.
This is because, in the inference stage,  a prediction can be made based on only the earlier predictions and the past history \textemdash{} information about
later predictions are not available. Hence, a mask needs to be placed here such that in the training stage we also do the same thing as in the inference stage.
The second sparse multi-head attention layer compares each element of the sequence in the decoder with the history sequence in the encoder} so that
we can learn from the past history. \cam{If non-stationarity happens, the comparison will tell the element is different from the historical elements that it is normally similar to, and therefore we should instead learn from other more similar historical elements, handling this non-stationarity. The following sparse feedforward layer further refines the encoding of each element.}

%% file: Evaluation.tex
\section{Evaluation}

In this section, we apply Forecaster to the problem of forecasting
taxi ride-hailing demand in Manhattan, New York City. We demonstrate
that Forecaster outperforms the state-of-the-art \ch{baselines (the Transformer
\cite{Transformer} and DCRNN \cite{DCRNN_ICLR2018}) and a conventional time series forecasting method (VAR \cite{VAR}).}

\subsection{Evaluation Settings}

\subsubsection{Dataset}

\begin{figure}[t]
\centering
\includegraphics[width=0.72\columnwidth]{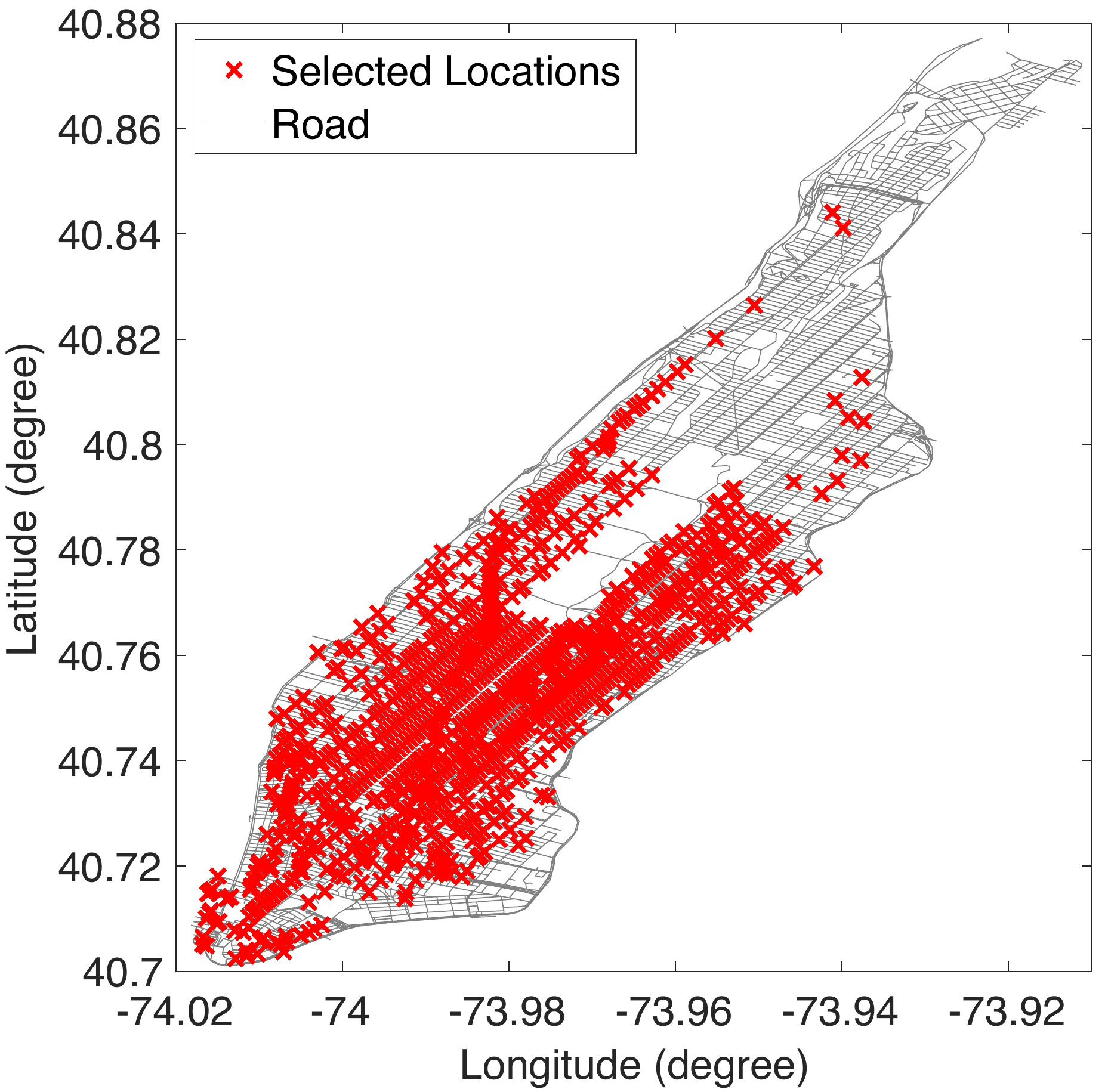}
\caption{\label{fig:Selected-locations}\ch{Selected locations in Manhattan.}}
\end{figure}

Our evaluation uses the NYC Taxi dataset \cite{NYC_Taxi} from 01/01/2009
to 06/30/2016 (7.5 years in total). This dataset records detailed
information for each taxi trip in New York City, including its pickup
and dropoff locations. Based on this dataset, we select 996 locations
with hot taxi ride-hailing demand in Manhattan of New York City, shown
in Figure \ref{fig:Selected-locations}. Specifically, we compute the taxi ride-hailing
demand at each location by accumulating the taxi ride closest to that location. Note that these selected locations
are not uniformly distributed, as different regions of Manhattan has
distinct taxi demand.\footnote{\ch{We use the following algorithm to select the locations. Our roadmap has 5464 locations initially. Then, we compute the average hourly taxi demand at each of these locations. After that, we use a threshold (= 10) and an iterative procedure to down select to the 996 hot locations. This algorithm selects the locations from higher to lower demand. Every time when a location is added to the pool of selected locations, we compute the average hourly taxi demand at each of the locations in the pool by remapping the taxi rides to these locations. If every location in the pool has a demand no less than the threshold, we will add the location; otherwise, remove it from the pool. We reiterate this procedure over all the 5464 locations. This procedure guarantees that all the selected locations have an average hourly taxi demand no less than the threshold.}} \ch{We compute the hourly taxi ride-hailing demand at these selected locations across time. As a result, our dataset contains 65.4 million data points in total (996 locations $\times$ number of hours in 7.5 years). As far as we know, it is the largest (in terms of data points) and longest (in terms of time length) dataset in similar types of study. Our dataset covers various types of scenarios and conditions (e.g., under extreme weather condition).} We split the dataset into three parts \textemdash{}
training set, validation set, and test set. Training set uses the
data in the time interval 01/01/2009 \textendash{} 12/31/2011 and 07/01/2012 \textendash{}
06/30/2015; validation set uses the data in 01/01/2012 \textendash{}
06/30/2012; and the test set uses the data in 07/01/2015 \textendash 06/30/2016.

Our evaluation uses hourly weather data from \cite{WeatherUnderground}
to construct (part of) the auxiliary information. Each record in this
weather data contains seven entries \textemdash{} temperature, wind
speed, precipitation, visibility, and the Booleans for rain, snow,
and fog.

\subsubsection{Details of the Forecasting Task}

In our evaluation, we forecast taxi demand for the next three hours  based
on the previous 674 hours and the corresponding auxiliary
information (i.e., \cam{use a history of four weeks around}; $T=674$,
$T'=3$ in Equation \eqref{eq:1}). Instead of directly inputing
this history sequence into the model, we first filter it. This filtering is based on the following observation: a
future taxi demand correlates more with the taxi demand at previous recent
hours, the similar hours of the past week, and the similar hours
on the same weekday in the past several weeks. In other words, we
shrink the history sequence and only input the elements relevant
to forecasting. Specifically, our filtered history
sequence contains the data for the following taxi demand (and the corresponding
auxiliary information):
\begin{itemize}\vspace{-4pt}
\item The recent past hours: $\mathbf{{x}}_{t-i},\;i=0,...,5$ ;
\item Similar hours of the past week: $\mathbf{{x}}_{t+i-j\times24},\;i=-1,...,5,\;j=1,..,6$
;
\item Similar hours on the same weekday of the past several weeks: $\mathbf{{x}}_{t+i-j\times24\times7},\;i=-1,...,5,\;j=1,..,4$.
\end{itemize}

\subsubsection{Evaluation Metrics}

Similar to prior work \cite{DCRNN_ICLR2018,STMGCN_AAAI2019}, we use
root mean square error (RMSE) and mean absolute percentage
error (MAPE) to evaluate the quality of the forecasting results. Suppose that
for the $j^{th}$ forecasting job ($j=1,\cdots,S$), the ground
truth is $\left\{ x_{t}^{i^{(j)}}\:|\:t=1,\cdots,T^{'},\:i=1,\cdots,N\right\} $,
and the prediction is $\left\{ \widehat{x_{t}^{i}}^{(j)}\:|\:t=1,\cdots,T^{'},\:i=1,\cdots,N\right\} $,
where $N$ is the number of locations, and $T'$ is the length of the
forecasted sequence. Then RMSE and MAPE are:
\begin{equation}
\begin{array}{r@{}l}
\mathrm{{RMSE}}&{}=\sqrt{\dfrac{1}{ST'N}\sum\limits_{j=1}^{S}\sum\limits_{t=1}^{T'}\sum\limits_{i=1}^{N}\left(\widehat{x_{t}^{i}}^{(j)}-x_{t}^{i^{(j)}}\right)^{2}}\\ \\
\mathrm{{MAPE}}&{}=\dfrac{1}{ST'N}\sum\limits_{j=1}^{S}\sum\limits_{t=1}^{T'}\sum\limits_{i=1}^{N}\left|\dfrac{\widehat{x_{t}^{i}}^{(j)}-x_{t}^{i^{(j)}}}{x_{t}^{i^{(j)}}}\right|
\end{array}
\end{equation}

Following practice in prior work \cite{STMGCN_AAAI2019}, we
set a threshold on $x_{t}^{i^{(j)}}$ when computing MAPE: if $x_{t}^{i^{(j)}}<10$,
disregard the term associated it. This practice prevents small $x_{t}^{i^{(j)}}$
dominating MAPE.

\subsection{Models Details}

We evaluate Forecaster and compare it against baseline models including \ch{VAR, DCRNN, and
the Transformer.}

\subsubsection{Our model: Forecaster}

Forecaster uses weather (\ch{7-dimensional vector}), weekday (one-hot encoding, \ch{7-dimensional vector}),
hour (one-hot encoding, \ch{24-dimensional vector}), and a Boolean for holidays (\ch{1-dimensional vector})
as auxiliary information (\ch{39-dimensional vector}). Concatenated with a spatial signal
(\ch{996-dimensional vector}), each element of the input sequence for Forecaster \ch{is a
1035-dimensional vector}. Forecaster uses one encoder layer and one decoder layer
(i.e., $N=1)$. Except for the sparse linear layer at the end \ch{of the decoder}, all
the layers of Forecaster use four neurons for encoding the data at
each location and 64 neurons for encoding the auxiliary information
and thus have 4048 neurons in total (i.e., $d_{model}=4\times996+64=4048$).
The sparse linear layer at the end has 996 neurons. Forecaster uses
the following loss function:
\begin{equation}
\cam{
\textrm{{loss}}\left(\cdot\right)=\eta\times\textrm{{RMSE}}^2+\textrm{{MAPE}}
}\end{equation}
where $\eta$ is a constant balancing the impact of $\textrm{{RMSE}}$
with $\textrm{{MAPE}}$, \cam{$\eta=8\times10^{-3}$}.

\subsubsection{\ch{Baseline model: Vector Autoregression}}
\ch{Vector autoregression (VAR) \cite{VAR} is a conventional multivariant
time series forecasting method. It predicts the future endogenous
variables (i.e., the spatial signal $\mathbf{{x}}_{t}$ in our case)
as a linear combination of the past endogenous variables and the current
exogenous variables (i.e., the auxiliary information $\mathbf{{a}}_{t}$
in our case):
\begin{equation}\label{eq:8}
\mathbf{{\hat{x}}}_{t+1}=A_{1}\mathbf{{x}}_{t}+\cdots+A_{p}\mathbf{{x}}_{t-p+1}+B\mathbf{{a}}_{t+1}
\end{equation}
where $\mathbf{{x}}_{t}\in\mathbb{{R}}^{N}$, $\mathbf{{a}}_{t+1}\in\mathbb{{R}}^{P}$,
$A_{i}\in\mathbb{{R}}^{N\times N},\;i=1,\ldots,p,$ $B\in\mathbb{{R}}^{N\times P}$.
Matrices $A_{i}$ and $B$ are estimated during the training stage. Our
implementation is based on Statsmodels\cite{statsmodels}, a standard Python package for statistics.}

\subsubsection{Baseline model: DCRNN}

DCRNN \cite{DCRNN_ICLR2018} \ch{is a deep learning model that} models the dependency relations between locations as a diffusion
process guided by a pre-defined distance metric. Then, it leverages
graph CNN to capture spatial dependency and RNN to capture the temporal
dependency within the data.

\subsubsection{Baseline model: Transformer}

The Transformer \cite{Transformer} uses the same input and loss function as Forecaster.
It also adopts a similar architecture except that all the layers are
fully-connected. For a comprehensive comparison, we evaluate
two versions of the Transformer:
\begin{itemize}
\item \ch{Transformer (same width)}: \ch{All the layers in this implementation have the same width as Forecaster. The linear layer at the end of decoder has a width of 996; other layers have a width of 4048 (i.e., $d_{model}=4048$).}
\item \ch{Transformer (best width)}: We vary the width of all the layers (except for
the linear layer at the end \ch{of decoder} which has a fixed width of 996) from 64 to 4096, and pick the best width
in performance to implement.
\end{itemize}

\subsection{Results}
\begin{figure}[b]
\centering
\includegraphics[width=0.7\columnwidth]{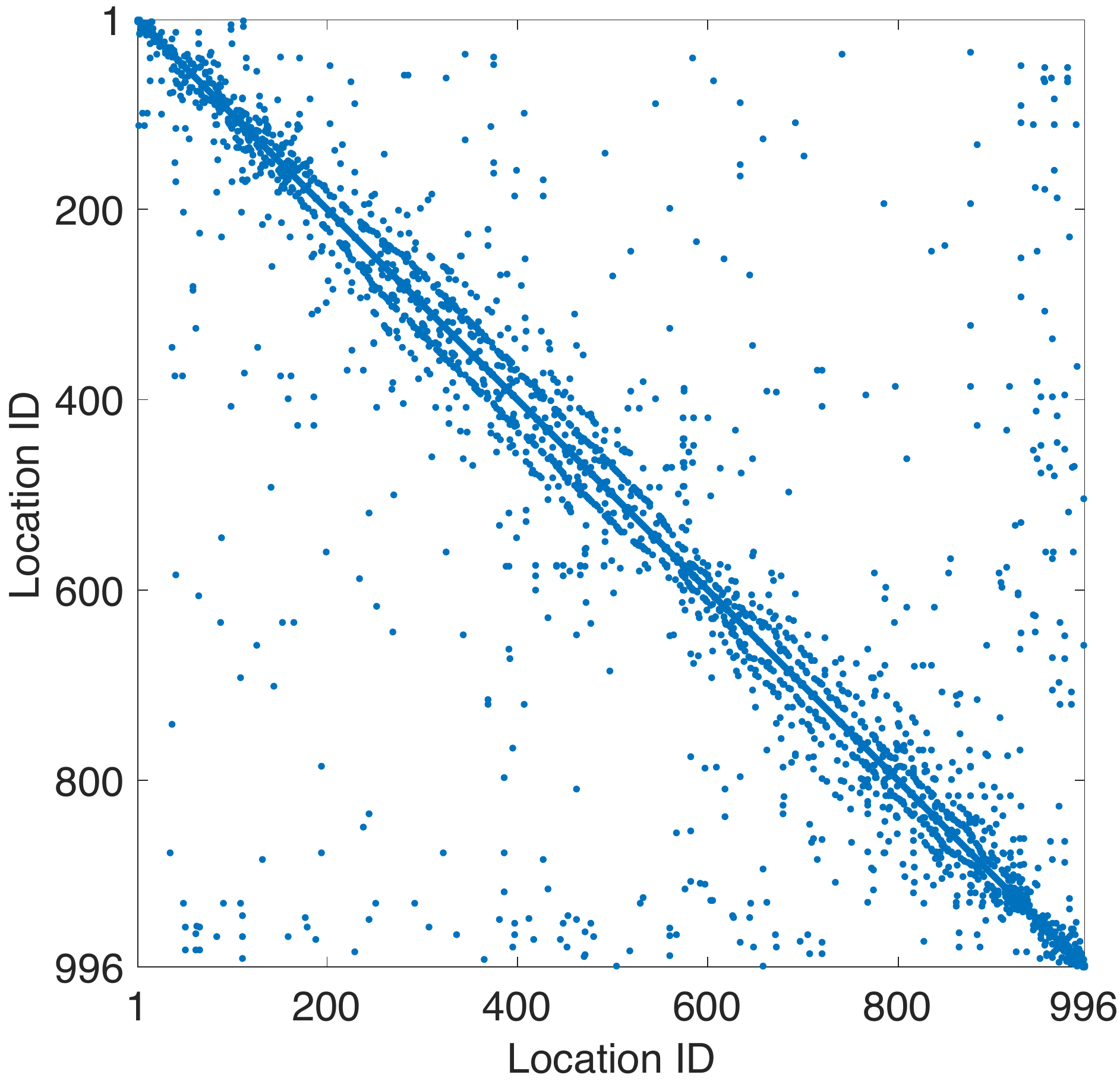}
\caption{\label{fig:Conditional-correlation-matrix}\ch{Structure of the conditional correlation
matrix (under a threshold of 0.1; each dot represents a non-zero entry).}}
\end{figure}

\begin{table*}
\caption{\label{tab:RMSE-and-MAPE}RMSE and MAPE of Forecaster and baseline models.}
\vspace{0.5em}
\centering
\begingroup
\renewcommand{\arraystretch}{1.2}
\begin{tabular}{|c|c|c|c|c|c|}
\hline 
Metrics & Model & Average & Next step & Second next step & Third next step\tabularnewline
\hline 
\hline 
 & \ch{VAR} & \ch{6.9991} & \ch{6.4243} & \ch{7.1906} & \ch{7.3476}\tabularnewline
\cline{2-6} \cline{3-6} \cline{4-6} \cline{5-6} \cline{6-6} 
 & DCRNN & 5.3750 \textpm{} 0.0691 & 5.1627 \textpm{} 0.0644 & 5.4018 \textpm{} 0.0673 & 5.5532 \textpm{} 0.0758\tabularnewline
\cline{2-6} \cline{3-6} \cline{4-6} \cline{5-6} \cline{6-6} 
RMSE & \ch{Transformer (same width)} & 5.6802 \textpm{} 0.0206 & 5.4055 \textpm{} 0.0109 & 5.6632 \textpm{} 0.0173 & 5.9584 \textpm{} 0.0478\tabularnewline
\cline{2-6} \cline{3-6} \cline{4-6} \cline{5-6} \cline{6-6} 
 & \ch{Transformer (best width)} & 5.6898 \textpm{} 0.0219 & 5.4066 \textpm{} 0.0302 & 5.6546 \textpm{} 0.0581 & 5.9926 \textpm{} 0.0472\tabularnewline
\cline{2-6} \cline{3-6} \cline{4-6} \cline{5-6} \cline{6-6} 
 & Forecaster & \textbf{5.1879 \textpm{} 0.0082} & \textbf{4.9629 \textpm{} 0.0102} & \textbf{5.2275 \textpm{} 0.0083} & \textbf{5.3651 \textpm{} 0.0065}\tabularnewline
\hline 
 & \ch{VAR} & \ch{33.7983} & \ch{31.9485} & \ch{34.5338} & \ch{34.9126}\tabularnewline
\cline{2-6} \cline{3-6} \cline{4-6} \cline{5-6} \cline{6-6} 
 & DCRNN & 24.9853 \textpm{} 0.1275 & 24.4747 \textpm{} 0.1342 & 25.0366 \textpm{} 0.1625 & 25.4424 \textpm{} 0.1238\tabularnewline
\cline{2-6} \cline{3-6} \cline{4-6} \cline{5-6} \cline{6-6} 
MAPE (\%) & \ch{Transformer (same width)} & 22.5787 \textpm{} 0.2153 & 21.8932 \textpm{} 0.2006 & 22.3830 \textpm{} 0.1943 & 23.4583 \textpm{} 0.2541\tabularnewline
\cline{2-6} \cline{3-6} \cline{4-6} \cline{5-6} \cline{6-6} 
 & \ch{Transformer (best width)} & 22.2793 \textpm{} 0.1810 & 21.4545 \textpm{} 0.0448 & 22.1954 \textpm{} 0.1792 & 23.1868 \textpm{} 0.3334\tabularnewline
\cline{2-6} \cline{3-6} \cline{4-6} \cline{5-6} \cline{6-6} 
 & Forecaster & \textbf{20.1362 \textpm{} 0.0316} & \textbf{19.8889 \textpm{} 0.0269} & \textbf{20.0954 \textpm{} 0.0299} & \textbf{20.4232 \textpm{} 0.0604}\tabularnewline
\hline
\end{tabular}
\endgroup
\vspace{1.5em}
\end{table*}

\begin{figure}
\centering
\includegraphics[width=0.75\columnwidth]{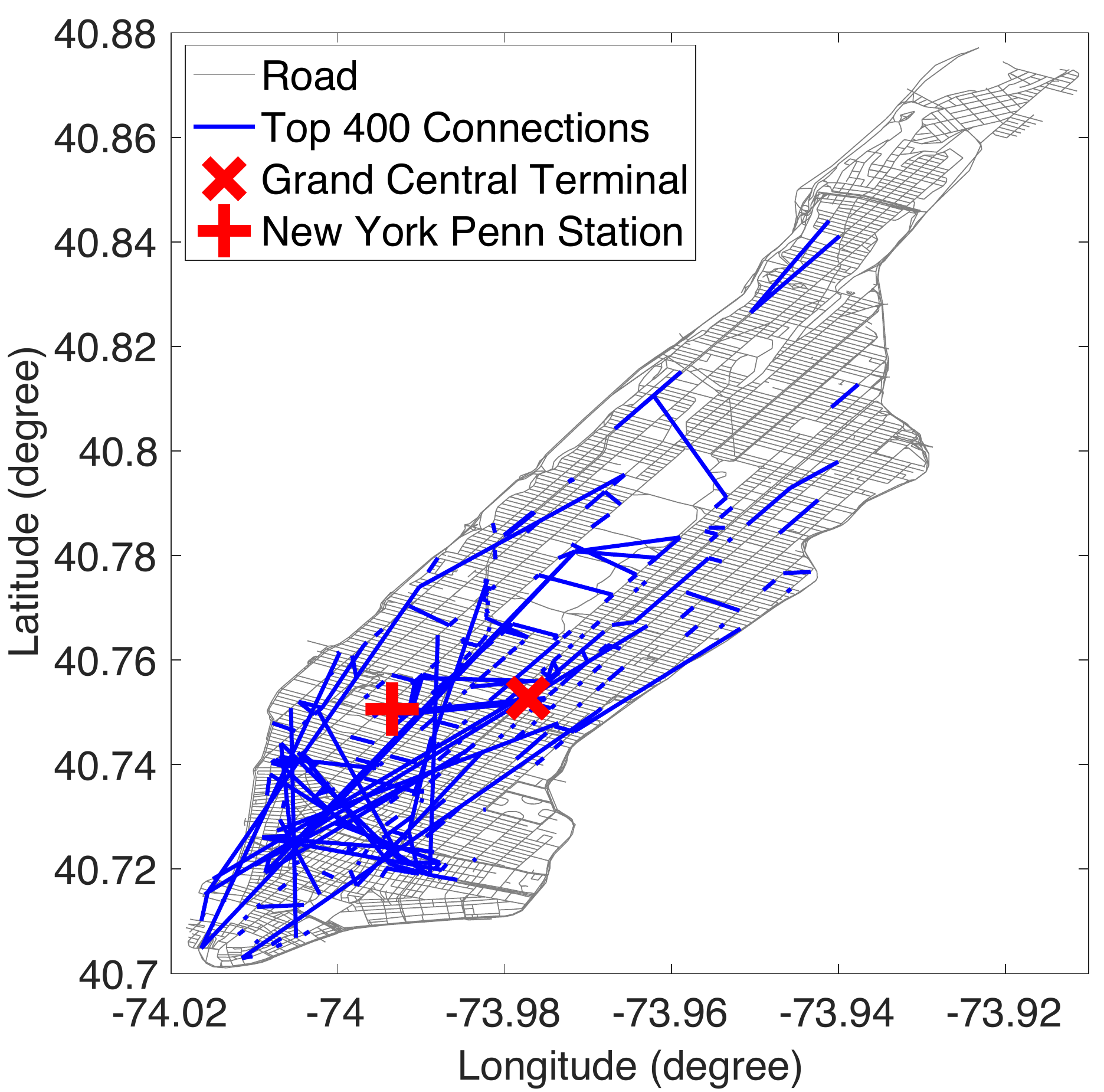}
\caption{\label{fig:Top-400-dependency}\ch{Top 400 dependency relations between
locations.}}
\end{figure}

Our evaluation of Forecaster starts by using Gaussian Markov random
fields to determine the spatial dependency between the data at different
locations. Based on the method in Section 2.2, we can obtain
a conditional correlation matrix where each entry of the matrix represents
the conditional correlation coefficient between two locations. If the absolute value of an entry is less than a threshold,
we will treat the corresponding two locations as conditionally independent,
and round the value of the entry to zero. \ch{This threshold can be chosen based only
on the performance on the validation set.} Figure \ref{fig:Conditional-correlation-matrix}
shows the structure of the conditional correlation matrix under a
threshold of 0.1. We can see that the matrix is sparse, which means
a location generally depends on just a few other locations other than
all the locations. We found that a location depends on only 2.5 other locations
on average. There are some locations which many other locations
depend on. For example, there is a location in Lower Manhattan
which 16 other locations depend on. This may be because there are
many locations with significant taxi demand in Lower Manhattan, with these
locations sharing a strong dependency. Figure \ref{fig:Top-400-dependency}
shows the top 400 spatial dependencies. We see
some long-range spatial dependency between remote locations. For
example, there is a strong dependency between Grand Central Terminal
and New York Penn Station, which are important stations in Manhattan
with a large traffic of passengers.

After determining the spatial dependency between locations, we use
the graph \ch{T}ransformer architecture of Forecaster to \ch{predict} the taxi
demand. Table~\ref{tab:RMSE-and-MAPE} contrasts the performance of Forecaster
to other baseline models. Here we run all the evaluated \ch{deep learning} models six times
(using different seeds) and report the mean and the standard deviation of the results. \ch{As VAR is not subject to the impact of random initialization, we run it once. We can see for all the evaluated models, the RMSE and MAPE of predicting the next step are lower than that of predicting later steps (e.g., the third next step). This is because, for all the models, the prediction of later steps is built upon the prediction of the next step, and thus the error of the former includes the error of the latter. Comparing the performance of these models, we can see the RMSE and MAPE of VAR is higher than that of the deep learning models. This is because VAR does not model well the non-linearity and non-stationarity within the data; it also does not consider the spatial dependencies between locations in the structure of its coefficient matrices (matrices $A_{i}$ and $B$ in Equation \eqref{eq:8}). Among the deep learning models, DCRNN and the Transformer perform similarly. The former captures the spatial dependency within the data but does not capture well the long-range temporal dependency, while the latter focuses on exploiting the long-range temporal dependency but neglects the spatial dependency. As for our method,}  Forecaster outperforms all the baseline
methods at every future step of forecasting. On average (over these
future steps), Forecaster achieves an RMSE of 5.1879 and a MAPE of
20.1362, which is 8.8210\% and 9.6192\% better than \ch{Transformer (best width)}, and 3.4809\% and 19.4078\% better than DCRNN. This
demonstrates the advantage of Forecaster \ch{in capturing both the spatial dependency and the long-range temporal dependency.}

%% file: RelatedWork.tex
\section{Related Work}

To our knowledge, this work is the first (1) to integrate
Gaussian Markov Random fields with deep learning to forecast
spatial and time-dependent data, using the former to derive a dependency graph; (2) to sparsify the architecture of the Transformer
based on the dependency graph, significantly improving the
forecasting quality of the result architecture. The most closely related work is a set of proposals
on forecasting spatial and time-dependent data and the Transformer, which we briefly
review in this section.

\subsection{Spatial and Time-Dependent Data Forecasting}

Conventional methods for forecasting spatial and time-dependent data such as ARIMA
and Kalman filtering-based methods \cite{liu2011,lippi2013} usually
impose strong stationary assumptions on the data, which are often violated \cite{DCRNN_ICLR2018}. Recently, deep learning-based methods
have been proposed to tackle the non-stationary and highly nonlinear nature
of the data \cite{DMVSTNET_AAAI2018,STResNet_AAAI17,STGCN_IJCAI2018,STMGCN_AAAI2019,Yao_AAAI2019,DCRNN_ICLR2018}.
Most of these works consist of two parts: modules to capture spatial
dependency and modules to capture temporal dependency. Regarding spatial dependency,
the literature mostly uses prior knowledge such as physical closeness between regions to derive an adjacency matrix and/or pre-defined distance/similarity metrics
to decide whether two locations are dependent or not. Then, based
on this information, they usually use a (standard or graph) CNN to characterize
the spatial dependency between dependent locations. However, these methods are not good predictors of dependency relations between the data at different locations. Regarding temporal dependency, available works \cite{DMVSTNET_AAAI2018,STGCN_IJCAI2018,STMGCN_AAAI2019,Yao_AAAI2019,DCRNN_ICLR2018}
usually use RNNs and CNNs to extract the long-range temporal
dependency. However, both RNN and CNN do not learn
well the long-range temporal dependency, with the number of operations
used to relate signals at two distant time positions in a sequence
growing at least logarithmically with the distance between them \cite{Transformer}.

We evaluate our architecture with the problem of forecasting taxi
ride-hailing demand around a large number of spatial locations. The
problem has two essential features: (1) These locations are not uniformly
distributed like pixels in an image, making standard CNN-based methods
\cite{DMVSTNET_AAAI2018,Yao_AAAI2019,STResNet_AAAI17} not good for this
problem; (2) it is desirable to perform multi-step forecasting, i.e., forecasting at several time instants in the future, this implying that the work
mainly designed for single-step forecasting \cite{STGCN_IJCAI2018,STMGCN_AAAI2019}
is less applicable. DCRNN \cite{DCRNN_ICLR2018} is the state-of-the-art
baseline satisfying both features. Hence, we compare our architecture
with DCRNN and show that our work outperforms DCRNN.

\subsection{Transformer}

The Transformer \cite{Transformer} avoids recurrence and instead
purely relies on the self-attention mechanism to let the data at distant positions
in a sequence to relate to each other directly. This benefits learning long-range temporal dependency. The \ch{Transformer
and its extensions have been shown to significantly outperform RNN-based methods in NLP
and image generation tasks \cite{Transformer,OpenAI_Transformer,Transformer_XL,XLNet,Bert,Image_Transformer,koncel}. 
It has also been applied to graph and node classification problems \cite{chen2019path, NIPS2019_Korea}.}
However, it is still unknown how to apply the architecture of Transformer
to spatial and time-dependent data, especially to deal with spatial dependency
between locations. Later work \cite{NonLocal_Net} extends the architecture
of Transformer to video generation. Even though this also needs to
address spatial dependency between pixels, the nature of the problem
is different from our task. In video generation, pixels exhibit
spatial dependency only over a short time interval, lasting for at most tens of frames \textemdash{}
two pixels may be dependent only for a few frames and become  \ch{independent} in later frames. On the contrary, in spatial and time-dependent data, locations
exhibit long-term spatial dependency lasting for months or even years.
This fundamental difference of the applications that we consider enables us to use Gaussian Markov random fields to determine the \ch{dependency graph} as
basis for sparsifying the Transformer. \ch{Child et al.}
\cite{Sparse_Transformer} propose another sparse Transformer architecture
with a different goal of accelerating the \ch{multi-head attention} operations
in the Transformer. This architecture is very different from our
architecture.

%% file: Conclusion.tex
\section{Conclusion}

Forecasting spatial and time-dependent data is challenging due to
complex spatial dependency, long-range temporal dependency, non-stationarity,
and heterogeneity within the data. This paper proposes Forecaster,
a graph Transformer architecture to tackle these challenges. Forecaster
uses Gaussian Markov random fields to determine the dependency
graph between the data at different locations. Then, Forecaster sparsifies the architecture of the Transformer based on the structure of the
graph and lets the sparsified Transformer (i.e., graph Transformer)
capture the spatiotemporal dependency, non-stationarity, and heterogeneity
in one shot. We apply Forecaster to the problem of forecasting taxi-ride
hailing demand at a large number of spatial locations. Evaluation results
demonstrate that Forecaster significantly outperforms state-of-the-art
baselines (the Transformer and DCRNN).
\ch{\ack We thank the reviewers. 
This work is partially supported by NSF CCF (award  1513936).}

%% file: Appendix.tex
\section*{\ch{Appendix: Multi-Head Attention}}
\ch{The multi-head attention layer is a core component of the Transformer
for capturing long-range temporal dependency within data.
It takes a query sequence $\left\{ \mathbf{{q}}_{t}\mid\mathbf{{q}}_{t}\in\mathbb{{R}}^{d_{model}},\:t=1,\ldots,T\right\} $,
a key sequence $\left\{ \mathbf{{k}}_{t}\mid\mathbf{{k}}_{t}\in\mathbb{{R}}^{d_{model}},\:t=1,\ldots,T\right\} $,
and a value sequence $\left\{ \mathbf{{v}}_{t}\mid\mathbf{{v}}_{t}\in\mathbb{{R}}^{d_{model}},\:t=1,\ldots,T\right\} $
as inputs, and outputs a new sequence $\left\{ \mathbf{{e}}_{t}\mid\mathbf{{e}}_{t}\in\mathbb{{R}}^{d_{model}},\:t=1,\ldots,T\right\} $
where each element of the output sequence is impacted by the corresponding query and all the keys and values, no matter how distant these keys and values are from the query in the temporal order, and thus captures the
long-range temporal dependency. The detailed procedure is as follows.}

\ch{First, the multi-head attention layer compares each query $\mathbf{q}_{t}$
with each key $\mathbf{k}_{j}$ to get their similarity
$\alpha_{tj}^{(h)}$ from multiple
perspectives (termed as \textit{multi-head}; $H$ is the number of heads, $h=1,\ldots,H$):
\begin{equation}
\begin{array}{rcl}
\alpha_{tj}^{(h)} & = & \mathrm{{softmax}}\left(\left\langle W_{(h)}^{Q}\mathbf{q}_{t},\:W_{(h)}^{K}\mathbf{k}_{j}\right\rangle \Big/\sqrt{\frac{d_{model}}{H}}\right)\\
\\
 \large& = & \dfrac{\mathrm{{exp}}\left(\left\langle W_{(h)}^{Q}\mathbf{q}_{t},\:W_{(h)}^{K}\mathbf{k}_{j}\right\rangle \Big/\sqrt{\frac{d_{model}}{H}}\right)}{\sum\limits_{i=1}^{T}\mathrm{{exp}}\left(\left\langle W_{(h)}^{Q}\mathbf{q}_{t},\:W_{(h)}^{K}\mathbf{k}_{i}\right\rangle \Big/\sqrt{\frac{d_{model}}{H}}\right)}
\end{array}\label{eq:9}
\end{equation}
where $\alpha_{ij}^{(h)}\in\left(0,\:1\right)$ is the similarity
between $\mathbf{q}_{i}$ and $\mathbf{k}_{j}$
under head $h$, $\sum_{j=1}^{T}\alpha_{ij}^{(h)}=1$; $W_{(h)}^{Q}, W_{(h)}^{K}\in\mathbb{{R}}^{\frac{d_{model}}{H}\times d_{model}}$
are parameter matrices for head $h$ that need to be learned; $\left\langle \cdot,\:\cdot\right\rangle $
is the inner product between two vectors.}

\cam{In our work, to balance the impact of spatial signals and
auxiliary information on the prediction, we first scale  $\mathbf{{q}}_{t}$
and then use its scaled version $\mathbf{{q}}_{t}^{\prime}$ instead
in Equation \eqref{eq:9} for computing the similarity $\alpha_{tj}^{(h)}$. 
Suppose in  $\mathbf{{q}}_{t}$ and $\mathbf{{k}}_{t}$, the first $d_{signal}$ dimensions are for encoding spatial signals, 
and the next $d_{aux}$ dimensions are for encoding auxiliary information, we compute $\mathbf{{q}}_{t}^{\prime}$ as:
\begin{equation}
\begin{array}{c}
\mathbf{{q}}_{t}^{\prime}=\mathbf{{r}}\circ\mathbf{{q}}_{t}\\
\\
\mathbf{{r}}=\left[\begin{array}{c}
\sqrt{\frac{1}{2}+\frac{d_{aux}}{2d_{signal}}}\cdot\mathbf{{1}}_{d_{signal}},\;\sqrt{\frac{1}{2}+\frac{d_{signal}}{2d_{aux}}}\cdot\mathbf{{1}}_{d_{aux}}\end{array}\right]^{{\rm {T}}}\\
\\
\mathbf{{1}}_{d}=\left[\begin{array}{ccc}
1 & \cdots & 1\end{array}\right]\in\mathbb{{R}}^{1\times d},\quad d=d_{signal}\;\textrm{{or}}\;d_{aux}
\end{array}
\end{equation}
where $\circ$ is the Hadamard product.} 

\ch{Second, the multi-head attention layer uses these similarities as
weights to generate a new sequence $\left\{ \mathbf{{e}}_{t}^{(h)}\mid\mathbf{{e}}_{t}^{(h)}\in\mathbb{{R}}^{d_{model}},\:t=1,\ldots,T\right\} $
for each head $h$:
\begin{equation}
\mathbf{{e}}_{t}^{(h)}=\sum_{j=1}^{T}\alpha_{tj}^{(h)}W_{(h)}^{V}\mathbf{v}_{j}
\end{equation}
where $W_{(h)}^{V}\in\mathbb{{R}}^{\frac{d_{model}}{H}\times d_{model}}$ is
another parameter matrix for head $h$ that needs to be learned.}

\ch{Third, the sequence under each head is concatenated and then used
to generate the final output sequence:
\begin{equation}
\mathbf{{e}}_{t}=W^{O}\left[\mathbf{{e}}_{t}^{(1)}\bigparallel\mathbf{{e}}_{t}^{(2)}\bigparallel\cdots\bigparallel\mathbf{{e}}_{t}^{(H)}\right]
\end{equation}
where $W^{O}\in\mathbb{{R}}^{d_{model}\times d_{model}}$ is a parameter matrix
that needs to be learned; $\cdot\bigparallel\cdot$ represents a concatenation
of two vectors.}

\ch{In summary, the multi-head attention layer needs to learn the parameter
matrices $W_{(h)}^{Q}$, $W_{(h)}^{K}$, $W_{(h)}^{V}$, $h=1,\cdots,H$, and $W^{O}$, which all can be treated as linear layers without
bias. In our architecture, we use sparse linear layers to replace
these linear layers, capturing the spatial dependency between locations.}